%% file: main.tex
\documentclass[acmsmall]{acmart}
\usepackage{siunitx}  
\usepackage{algorithmic}
\usepackage{algorithm}

\newcommand{\e}{\mathrm{e}}
\newcommand{\revision}{}

\setcopyright{acmcopyright}
\copyrightyear{0000}
\acmYear{0000}
\acmDOI{XXXXXXX.XXXXXXX}

\acmJournal{TECS}
\acmVolume{00}
\acmNumber{0}
\acmArticle{000}
\acmMonth{0}

\acmSubmissionID{64}

\begin{document}

\title{Hardware-Friendly Delayed Feedback Reservoir for Multivariate Time Series Classification}


\author{Sosei Ikeda}
\affiliation{%
  \institution{Kyoto University}
  \city{Kyoto}
  \country{Japan}}
\email{sikeda@easter.kuee.kyoto-u.ac.jp}

\author{Hiromitsu Awano}
\affiliation{%
  \institution{Kyoto University}
  \city{Kyoto}
  \country{Japan}} 
\email{awano@i.kyoto-u.ac.jp}

\author{Takashi Sato}
\affiliation{%
  \institution{Kyoto University}
  \city{Kyoto}
  \country{Japan}}
\email{takashi@i.kyoto-u.ac.jp}

\renewcommand{\shortauthors}{Ikeda et al.}

\begin{abstract}
\input{sub/abst.input_tex}

\end{abstract}



\received{00 January 0000}
\received[revised]{00 January 0000}
\received[accepted]{00 January 0000}
\maketitle

\input{sub/intro.input_tex}
\input{sub/reservoir.input_tex}
\input{sub/r.input_tex}
\input{sub/propose.input_tex}
\input{sub/evaluation.input_tex}
\input{sub/conclusion.input_tex}
\input{sub/ack.input_tex}

\bibliographystyle{ACM-Reference-Format}
\bibliography{ref}

\end{document}

%% file: sub/abst.input_tex
%
Reservoir computing is attracting attention as a machine learning technique for edge computing.
In time-series classification tasks,
the number of features obtained using a reservoir depends on the length of the input series.
Therefore, the features must be converted to a constant-length intermediate representation (IR),
such that they can be processed by an output layer.
Existing conversion methods involve computationally expensive matrix
inversion that significantly increases the circuit size and
requires processing power when implemented in hardware.
In this paper, we propose a simple but effective
IR, namely dot-product-based reservoir representation (DPRR), for reservoir computing
based on the dot product of data features.
Additionally, we propose a hardware-friendly
delayed feedback reservoir (DFR) consisting of
a nonlinear element and delayed feedback loop with DPRR.
The proposed DFR successfully classified multivariate time series data
that has been considered particularly difficult to implement efficiently in hardware.
In contrast to conventional DFR models that require analog circuits,
the proposed model can be implemented in a fully digital manner suitable for high-level syntheses.
A comparison with existing machine learning methods
via field-programmable gate array implementation
using 12 multivariate time-series classification tasks
confirmed the superior accuracy and small circuit size of the proposed method.

%% file: sub/intro.input_tex

\section{Introduction}
\label{sec:intro}

Edge computing,
which is gaining attention in today's Internet of Things (IoT) society,
is a method of processing and exchanging information with other devices in
the vicinity of data sources~\cite{shi2016edge}.
To reduce the energy consumption of edge devices,
processing near data sources such as sensors is crucial.
In general, data sources generate a large amount of time-series data that
must be processed in real time.
In addition, because
processing is performed at the edge devices,
they must operate with a low power consumption
and be implemented using small hardware resources.

Reservoir computing is a machine learning method
that uses a reservoir.
Input signals are transformed nonlinearly to a higher-dimensional space via
a layer called a reservoir with fixed weights,
and only the weights of the output layer are learned~\cite{lukovsevivcius2009reservoir}.
This method is considered to be particularly suitable for edge computing
because it is capable of processing time-series data
with its recurrent structure that reflects past inputs.
In addition, the reservoir layer does not require learning and can therefore be implemented
in hardware utilizing a variety of physical phenomena~\cite{tanaka2019recent}.

The greatest obstacle in applying reservoir computing to
real-world time-series classification tasks is the transformation of features
obtained via the reservoir~\cite{chen2013model}.
In general, the number of features obtained by the reservoir varies
with the length of the input data series,
thereby preventing the structure of the output layer from being fixed.
Therefore, it is necessary to convert these features to an intermediate representation (IR)
that is independent of the data length, such that
they can be processed using an output layer with a fixed structure.
Thus far, no established method has been proposed for this conversion.

For echo state networks (ESNs)~\cite{jaeger2001echo} that are another
variant of reservoir computing, several transformation methods have
been proposed~\cite{pascanu2015malware,aswolinskiy2016time}.  However, those
used in existing studies exhibited less classification accuracy
than existing neural-network-based classification methods,
especially in multivariate time-series classification tasks~\cite{bianchi2020reservoir}.
The reservoir model space (RMS)~\cite{bianchi2020reservoir} is an example of IR that facilitates
the classification of multivariate time-series data and improves the classification accuracy.
However, converting features to RMS involves the inversion of a large matrix,
which requires a substantial amount of hardware resources and computation time.

In this paper, we propose
a new feature transformation method that achieves both
high accuracy and small circuit size.
A dot-product-based reservoir representation (DPRR) is devised
by focusing on the convolution of the time development of the features.
We evaluated the effectiveness of applying DPRR
to a delayed feedback reservoir (DFR)~\cite{appeltant2011information},
which is a method of reservoir computing that enables a particularly compact construct
consisting of only one nonlinear element and a delayed feedback loop.
The combination of DPRR with DFR demonstrated excellent classification accuracy
on multivariate time-series classification tasks
and the smallest circuit size compared to existing machine learning classifiers based on neural networks.

The important contributions of this work can be summarized as follows:
\begin{enumerate}
 \item A novel IR called the dot-product-based reservoir representation (DPRR),
       which is computationally efficient and effective
       for improving the classification accuracy of multivariate time series data, is proposed.
 \item A fully digital delayed feedback reservoir (DFR) model with DPRR is proposed.
       The accuracy of DFR on multivariate time-series classification tasks
       is thoroughly evaluated. 
       The accuracy and hardware cost are compared with
       those of existing machine learning classifiers.
 \item A construct of input-data masking that supports multivariate time-series inputs
       and suppresses inference accuracy variation is defined.
\end{enumerate}

The remainder of this paper is organized as follows. In
Section~\ref{sec:reservoir}, the concept and operation of
reservoir computing are explained.  In
Section~\ref{sec:r}, existing reservoir representations are
briefly reviewed.
In Section~\ref{sec:propose}, a reservoir representation called
DPRR and a new DFR model that uses DPRR are proposed. 
In Section~\ref{sec:evaluation}, an evaluation of the effectiveness of DPRR
and classification accuracy of the proposed DFR implemented
on a field-programmable gate array (FPGA), are presented.
Finally, in Section~\ref{sec:conclusion}, we conclude the paper.

%% file: sub/reservoir.input_tex

\section{Reservoir Computing}
\label{sec:reservoir}

This section explains the concept of reservoir computing.
Subsequently, ESN and DFR that are typical implementations of reservoir computing are discussed.

\subsection{Concept of Reservoir Computing}
\label{subsec:reservoir}

Reservoir computing is a machine learning method comprising three major components or layers:
input layer, reservoir, and output layer~\cite{lukovsevivcius2009reservoir}.
First, the input layer converts the input signal into a format suitable for entering the reservoir layer.
Subsequently, the reservoir transforms the signal into a high-dimensional nonlinear vector.
Throughout this paper, we refer to this vector as the reservoir state
and to its components as features.
Finally, the output layer performs pattern recognition using a simple learning algorithm.
The characteristics of each layer are defined by the weights that connect the nodes.
The weights of the input layer and reservoir are fixed and unaltered,
whereas that of the output layer is determined through training.

In time-series processing, reservoir computing is useful for two main tasks:
regression and classification~\cite{schrauwen2007overview}.
In the regression tasks, one output is expected for each input.
In classification tasks, a single output is expected from multiple time-series inputs.
The number of features obtained by the reservoir layer depends on the length of the input series;
thus, the features need
to be converted into an intermediate representation
of a fixed length, regardless of the length of the inputs,
such that they can be processed at the output layer of a fixed size~\cite{chen2013model}.
In this paper, the intermediate representation is called the {\em reservoir representation}.
In the following,
for the sake of simplicity, we assume that the processing tasks are classification tasks
and consider a model that involves conversion
to a reservoir representation in its operations.

\subsection{Echo State Network (ESN)}
\label{subsec:ESN}

\begin{figure}[tbh]
 \centering
 \includegraphics[width=1.0\linewidth]{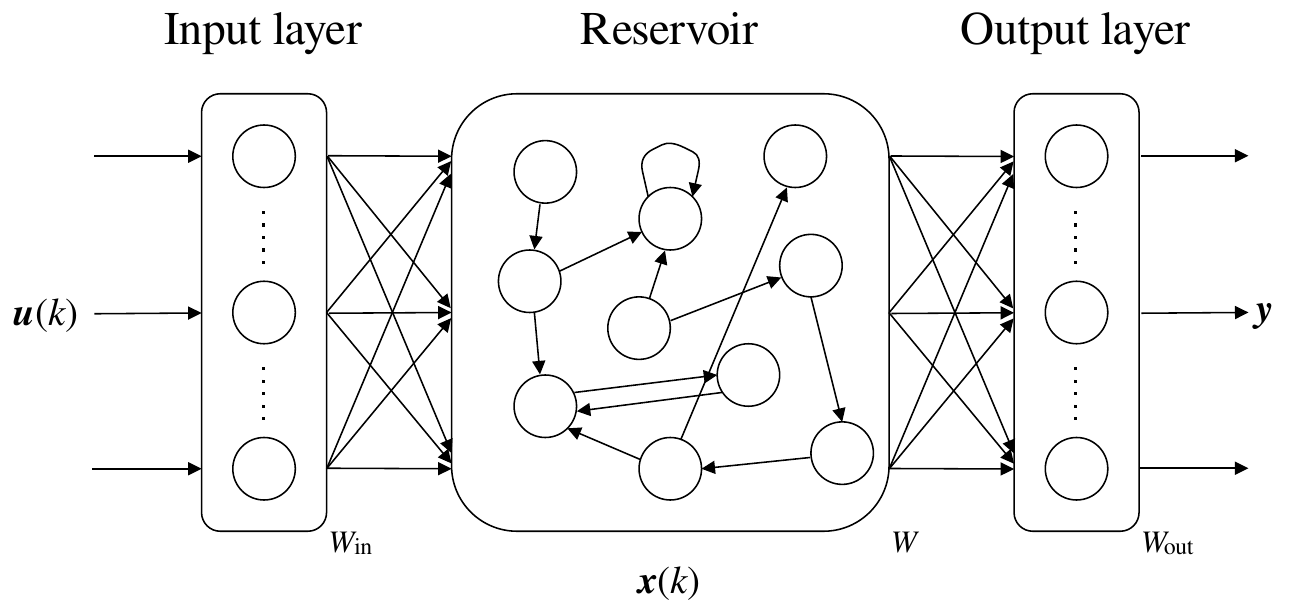}
 \caption{Conceptual diagram of ESN~\cite{tanaka2019recent}.
 \revision{The reservoir in the ESN is a recurrent neural network with fixed weight $W$.}}
 \label{fig:ESN}
\end{figure}

An ESN is a realization of reservoir computing.
The reservoir can be considered as
a recurrent neural network with fixed weights.
Figure~\ref{fig:ESN} depicts a
conceptual diagram of the ESN.

First, the input time series is represented as a series of vectors
$\boldsymbol{u}(k) \in\mathbb{R}^{N_u} \; (k=1, 2, \ldots, T)$.
Each $\boldsymbol{u}(k)$ represents a vector of real numbers,
$T$ is the length of the series or number of sampling time steps,
and $N_u$ denotes the number of variables.
When the number of nodes in the reservoir is $N_x$,
the weights of the input layer are represented as a matrix $W_{\mathrm{in}}$
with $N_x$ rows and $N_u$ columns.
The weights of the reservoir are represented as a matrix $W$
with $N_x$ rows and $N_x$ columns,
and the hyperbolic tangent function is used as an activation function.
The time evolution of the reservoir state $\boldsymbol{x}(k) \in\mathbb{R}^{N_x}$ is expressed as follows:
\begin{equation}
 \boldsymbol{x}(k+1)=\mathrm{tanh}(W_{\mathrm{in}}\boldsymbol{u}(k+1)+W\boldsymbol{x}(k)).
 \label{inputu}
\end{equation}

Next, the reservoir representation $\boldsymbol{r}\in\mathbb{R}^{N_r}$
is obtained from the reservoir states $\boldsymbol{x}(1),
\boldsymbol{x}(2), \ldots, \boldsymbol{x}(T)$.
In most cases, even within a dataset, the length
of the input time series varies for each input data instance. For example, in spoken language
classification, the lengths of the time series of the input
corresponding to "1" and "14" differ because of the difference in the
number of syllables. 
Accordingly, the length of the reservoir state $T$ differs.
Reservoir representations should be able to absorb these differences.
Regardless of the length of the input series, a constant-length vector should always be produced.
This is a strict requirement that comes from the fact that the output
layer that converts the reservoir representation to output
$\boldsymbol{y}$, is fixed~\cite{chen2013model}.
Here, the weight of the output layer, $W_{\mathrm{out}}$,
is trained in advance for the use of the ESN,
such that $\boldsymbol{y}=W_{\mathrm{out}}\boldsymbol{r}$
provides a good prediction.
Here, $\boldsymbol{y}\in\mathbb{R}^{N_y}$ is the output and $N_y$ is the number of classification classes.
Note that the target output is given in a one-hot representation,
with 1 for the correct label and 0 for the others.

\subsection{Delayed Feedback Reservoir (DFR)}
\label{subsec:DFR}

\begin{figure}[tbh]
 \centering
 \includegraphics[width=1.0\linewidth]{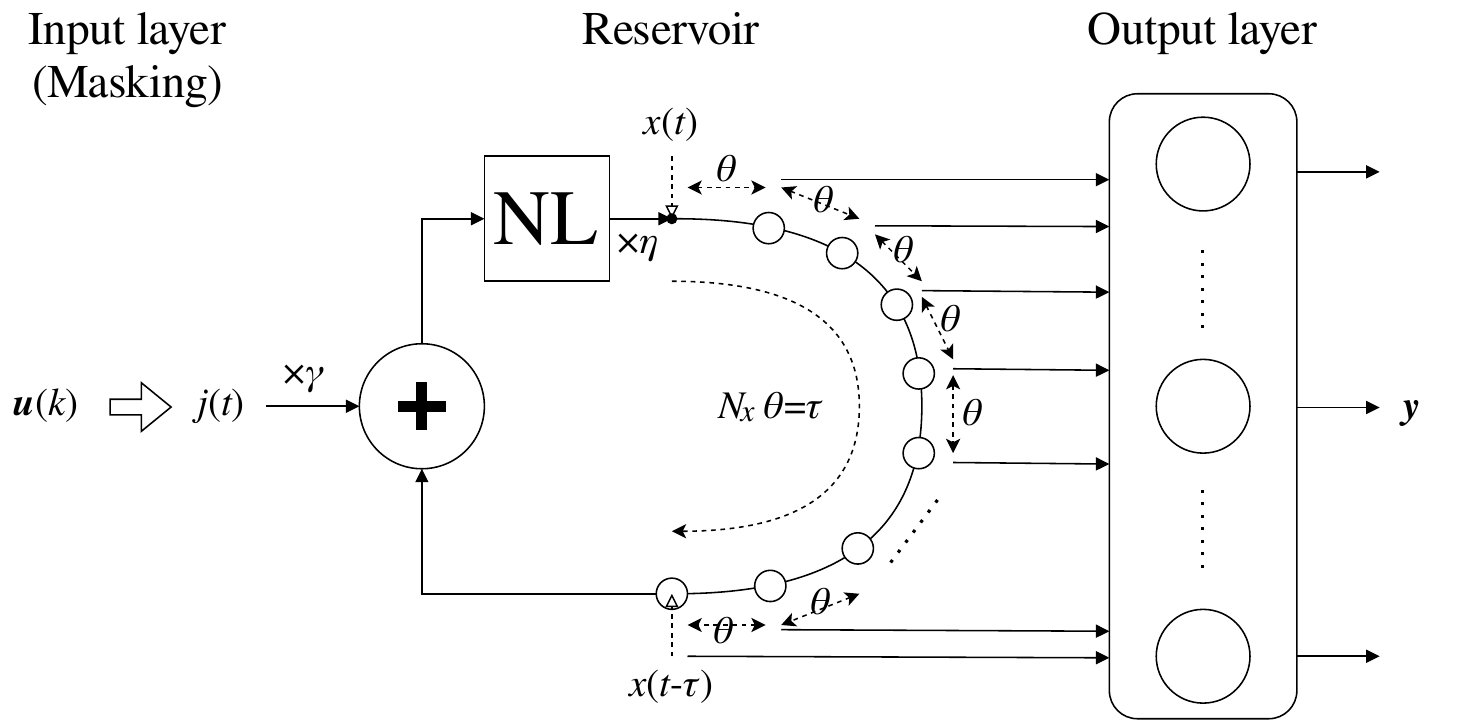}
 \caption{Conceptual diagram of DFR~\cite{appeltant2011information}.
 \revision{The reservoir in the DFR consists of a nonlinear element (NL)
 and a feedback loop with a total delay $\tau$.
 The feedback loop consists of $N_x$ virtual nodes having
 equal time intervals $\theta$.}}
 \label{fig:DFR}
\end{figure}

The DFR
is another realization of reservoir computing
in which the reservoir consists of only one nonlinear element and a delayed feedback loop.
Compared with other reservoir computing schemes,
DFR is easier to implement as hardware because of its simple structure~\cite{tanaka2019recent}.
Figure~\ref{fig:DFR} illustrates a conceptual diagram of DFR.

Most of the existing hardware implementations of the reservoir layer in DFR~\cite{appeltant2011information,soriano2014delay}
operate entirely using analog circuits.
Therefore, the input $\boldsymbol{u}(k)$,
typically represented as digital values such as the IEEE-754 float, is
converted to an analog signal before passing to the reservoir layer.
The analog output of the reservoir layer is then converted into a digital value.
More specifically, after
applying the masking process to the input time series $\boldsymbol{u}(k)$, it is
converted to analog signal $j(t)$ (Fig.~\ref{fig:DFR}).
This masking is a preprocessing step for the input signal, which is
equivalent to multiplying the input by $W_{\mathrm{in}}$ in the case of ESN.
\begin{figure}[tbh]
 \centering
 \includegraphics[width=1.0\linewidth]{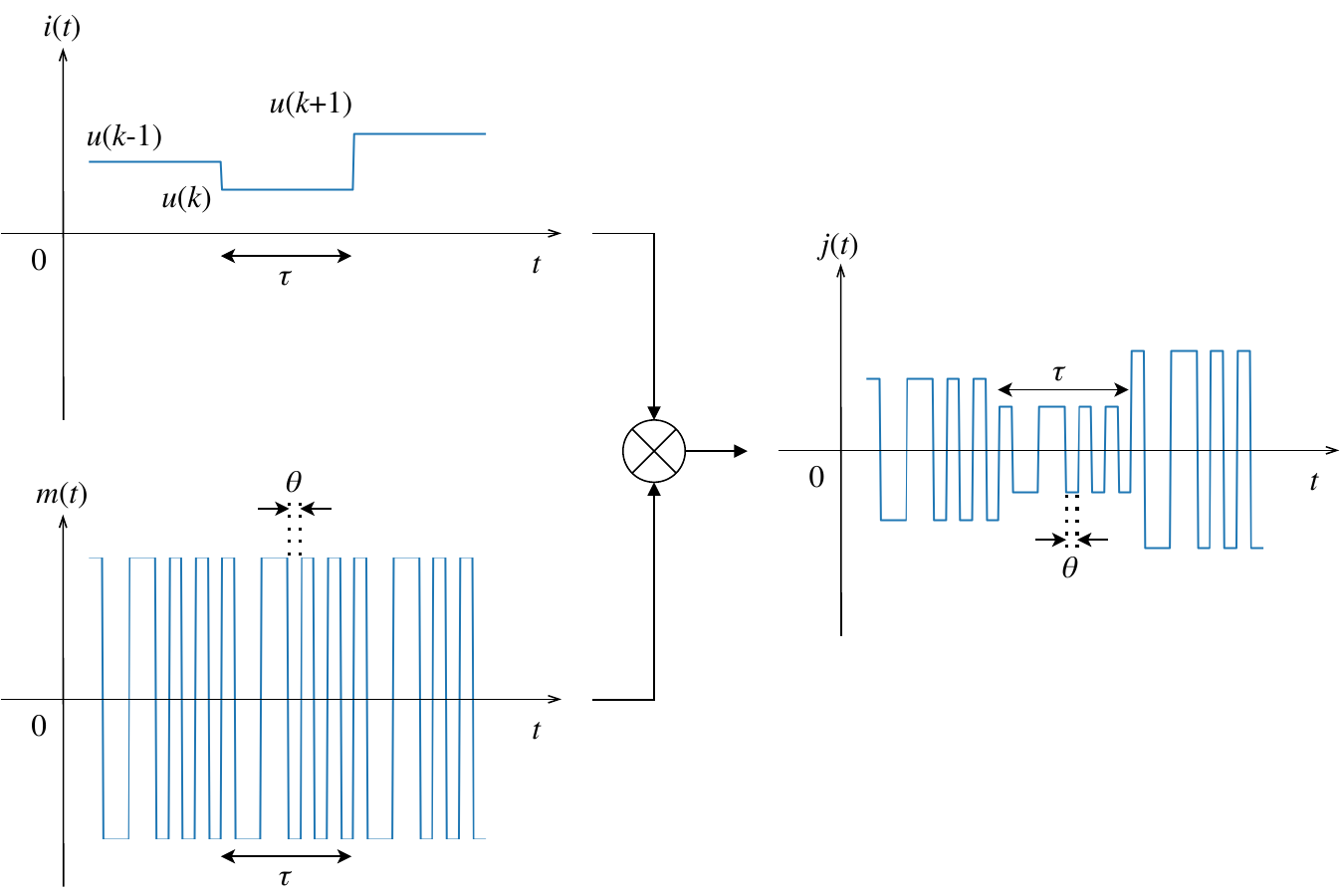}
 \caption{Masking process (univariate input example for clarity).
 \revision{$i$ is the output of digital-to-analog conversion of the original digital input signal $u$.
 The signal $i$ maintains a constant value for each time duration of $\tau$.
 The mask signal $m$ changes at time intervals of $\theta$ and has a period of $\tau$.
 The input to the reservoir is given as $j(t) = i(t) \cdot m(t)$.
 }}
 \label{fig:mask}
\end{figure}

The purpose of masking is to modulate the input signal using the mask
signal.  Figure~\ref{fig:mask} illustrates an example of mask processing for
a univariate time-series input.  The input signal $i(t)$ is
multiplied by the mask signal with a faster sample rate and
amplitude of either $+1$ or $-1$.
Note that the masking process for univariate inputs
has been explained in several studies~\cite{appeltant2011information,appeltant2014constructing};
however, that for the multivariate input case
has not yet been adequately investigated.

The masking process for a multivariate input signal is defined as follows.
Let $i_{1}(t), i_{2}(t), \ldots, i_{N_{u}}(t)$ be
the multivariate input signals:
\begin{equation}
 \left(
 \begin{array}{c}
 i_{1}(t) \\
 i_{2}(t) \\
 \cdots \\
 i_{N_{u}}(t) \\
 \end{array}
 \right)
 =\boldsymbol{u}(k).
\end{equation}
Here, the input $\boldsymbol{u}(k)$ remains the same for the time duration $\tau$.
The length of the mask patterns is equal to the virtual nodes $N_x$
in the DFR, which is designed such that
the following relationship holds:
\begin{equation}
 N_x\theta =\tau.
\end{equation}
Here, the delay between the nodes in the reservoir is $\theta$.
Therefore, the constant input during $\tau$ is divided into $N_x$
equal intervals of $\theta$, and each mask value is applied for the time
interval of $\theta$.
The converted input $j(t)$ can be expressed as follows:
\begin{equation}
 j(t)=\sum_{a=1}^{N_u}m_{a}(t)i_{a}(t),
\end{equation}
where $m_{1}(t), m_{2}(t), \ldots, m_{N_{u}}(t)$ are mask functions, each
having a length of $N_x$.

Once $j(t)$ is obtained, $j(t)$ is multiplied by $\gamma$ and added to
the feedback signal in the reservoir.  It is then entered into a
one-input, one-output nonlinear (NL) device, whose output is multiplied by
$\eta$ and becomes $x(t)$.
Thereafter, $x(t)$ is added to $\gamma j(t)$ through a
feedback loop with a total delay time of $\tau$.
Virtual nodes with interval $\theta$ are serially connected
to form a feedback loop in the reservoir. These nodes
operate as shift registers.  The values stored in the virtual nodes
are the features and they collectively form a vector with $N_x$ elements,
which is the reservoir state. Therefore, the state that this reservoir
holds in the delay elements is expressed as:
\begin{equation}
 \boldsymbol{x}(k)\equiv[x(k\tau -\theta ), x(k\tau -2\theta ), \ldots , x(k\tau -\tau)].
\end{equation}

Most of the delayed feedback reservoirs use the Mackey-Glass model~\cite{mackey1977oscillation}
as a nonlinear element\cite{appeltant2011information,soriano2014delay,alomar2015digital},
which is expressed as follows:
\begin{equation}
 \frac{\mathrm d}{{\mathrm d}t}x(t)=-x(t)+\frac{\eta [x(t-\tau)+\gamma j(t)]}{1+[x(t-\tau)+\gamma j(t)]^p},
  \label{eq:MG}
\end{equation}
where $p$ is an adjustable parameter. 
\revision{There are two reasons why this model is often used~\cite{appeltant2011information}.
First, it is easy to implement in analog electronic circuits.
Second, the nonlinearity can be adjusted by changing $p$.}
The existing studies often use $p=7$
for ease of hardware implementation.

Next, the reservoir representation $\boldsymbol{r}\in\mathbb{R}^{N_r}$
is obtained for the reservoir states $\boldsymbol{x}(1),
\boldsymbol{x}(2), \ldots, \boldsymbol{x}(T)$ for use in the output
layer. The construction of a reservoir representation is critically important
because it significantly affects both the accuracy of the DFR and
hardware resource usage.  We discuss the implementation
of the reservoir representation in the next section. The construct
and function of the output layer are similar to those of ESN. The weight of
the output layer, $W_{\mathrm{out}}$ is learned in advance, as
in ESN.

%% file: sub/r.input_tex

\section{Reservoir Representation}
\label{sec:r}

The reservoir maps the input signal to a higher dimensional space~\cite{schrauwen2007overview}.
Let the elements of $\boldsymbol{x}(k)$ (i.e., the features) be $[x(k)_1, x(k)_2, \ldots, x(k)_{N_x}]$.
The reservoir then yields a total of $T\cdot N_x$ features
that are denoted by $x(k)_n (k=1,2,\ldots,T;\; n=1,2,\ldots,N_x)$.
As explained in the previous sections, because
the length of the input series $T$ may vary,
so does the number of features.
In this situation, the features cannot be processed with an output layer of a fixed size.
Consequently, conversion to reservoir representation with a fixed length is required.
The following subsections review the existing reservoir representations.

\subsection{Last Reservoir State (LRS)}
\label{subsec:last}

In \cite{pascanu2015malware}, LRS after
all the input series were entered was used as the reservoir representation.
The reservoir representation is then expressed as:
\begin{equation}
 \boldsymbol{r}\equiv\boldsymbol{x}(T).
\end{equation}
The underlying assumption of this method is that the reservoir cumulatively
stores information about past inputs.
Therefore, the LRS is considered to contain all information for the
entire input.

The number of dimensions available with this representation
is $N_x$ and it does not depend on the series length $T$.  However,
$N_x$ is relatively small (otherwise, the hardware cost increases).  As a
result, the prediction of reservoirs using this reservoir representation is less
accurate than that using existing neural-network-based classifiers,
particularly for multivariate time-series classification
tasks~\cite{bianchi2020reservoir}.

\subsection{Direct Reservoir State (DRS)}
\label{subsec:mode}

In the DRS,
the reservoir state
at each sample time is used as a reservoir
representation:
\begin{equation}
 \boldsymbol{r}(k)\equiv\boldsymbol{x}(k).
\end{equation}
As the reservoir representation is available for every sample time $T$,
the outputs, $\boldsymbol{y}(1), \boldsymbol{y}(2), \ldots,
\boldsymbol{y}(T)$, are calculated using the output layer.
The mode of the $T$-inferred labels 
is used as the output inference label.

This representation is often used for time-series
classification using
DFR~\cite{appeltant2011information,soriano2014delay}.
Similar to the LRS, the number of features in DRS is $N_x$.
Owing to its low
dimensionality, this method also yields less accurate results than
existing neural-network-based classifiers, especially for 
multivariate time series classification
tasks~\cite{bianchi2020reservoir}.

\subsection{Maximal Reservoir States (MRSs)}
\label{subsec:all}

In the MRSs, $T_{\mathrm{max}}$, which is the maximum
series length $T$ of all possible inputs, is applied to all other
inputs to equalize the feature length.
$T_{\mathrm{max}}\cdot N_x$
features denoted by $x(k)_n (k=1,2,\ldots,T_{\mathrm{max}};\;
n=1,2,\ldots,N_x)$ are then used as a reservoir
representation~\cite{cabessa2021efficient}. In this method, there are
several options for extending short data to the maximum length.  The
first option is to append zeros at the end of
the input for which no value is given:
\begin{equation}
 \label{eq:all_upad}
 \boldsymbol{u}(k)\equiv\textbf{0}\; (k=T+1,T+2,\ldots,T_{\mathrm{max}}).
\end{equation}
The other option is to append zeros for the reservoir state:
\begin{equation}
 \label{eq:all_xpad}
 \boldsymbol{x}(k)\equiv\textbf{0}\; (k=T+1,T+2,\ldots,T_{\mathrm{max}}).
\end{equation}

The greatest drawback of this representation is the difficulty in
determining the necessary and sufficient $T_{\mathrm{max}}$ value in advance.
Setting $T_{\mathrm{max}}$ to be too large would increase hardware resources.
In addition, the zero-filling process of the input
(Eq.~(\ref{eq:all_upad})) requires additional processing until
$T_{\mathrm{max}}$ is reached, which reduces throughput,
whereas the zero-filling required for the reservoir state
(Eq.~(\ref{eq:all_xpad}))
cannot add additional information.

\subsection{Output Model Space (OMS)}
\label{subsec:oms}

In the OMS, linear regression is first used to predict the input
that is one time ahead
of the reservoir state:
\begin{equation}
 \label{eq:oms1}
 \boldsymbol{u}(k+1)=R_{\mathrm{oms}}\boldsymbol{x}(k)+\boldsymbol{r}_{\mathrm{oms}}.
\end{equation}
Here, $R_{\mathrm{oms}}$ and $\boldsymbol{r}_{\mathrm{oms}}$ are
constant regardless of $k$ and are subject to regression.
$\boldsymbol{x}'(k)$ is defined as:
\begin {equation}
 \boldsymbol{x}'(k)\equiv[\boldsymbol{x}(k),1].
\end{equation}
Eq.~(\ref{eq:oms1}) can be rewritten as:
\begin{equation}
 \boldsymbol{u}(k+1)=R'_{\mathrm{oms}}\boldsymbol{x}'(k).
\end{equation}
Note that the dimension of
$R'_{\mathrm{oms}}\in\mathbb{R}^{N_u\times(N_x+1)}$ holds.  Let
\begin{eqnarray}
 U^+ &\equiv [\boldsymbol{u}(1), \boldsymbol{u}(2), \ldots, \boldsymbol{u}(T)] \ \mathrm{and} \\
 X'  &\equiv [\boldsymbol{x}'(0), \boldsymbol{x}'(1), \ldots, \boldsymbol{x}'(T-1)].
\end{eqnarray}
Then, $R'_{\mathrm{oms}}$ may be ridge regressed
using the Moore-Penrose pseudoinverse with
$E$ as the $(N_x+1)$th-order unit matrix and $\lambda$ as a parameter:
\begin{equation}
\label{eq:oms2}
R'_{\mathrm{oms}}=U^+X'^T(X'X'^T+\lambda E)^{-1}.
\end{equation}
OMS is a single-column rearrangement of $R'_{\mathrm{oms}}$~\cite{chen2013model,bianchi2020reservoir}:
\begin{equation}
 \boldsymbol{r} \equiv\mathrm{vec}(R'_{\mathrm{oms}}).
\end{equation}
The derivation of OMS is based on the idea that there is a constant linear transformation
from the reservoir state to the input that is
one {time ahead,
because ESNs with a linear output layer are
effective for time series prediction tasks~\cite{chen2013model}.
In OMS, the number of features is $N_u\cdot (N_x+1)$,
which is independent of the series length $T$
and shows better classification accuracy than LRS and MRS~\cite{bianchi2020reservoir}.
However, OMS involves the computation of a high-dimensional inverse matrix,
which increases the circuit size when implemented in hardware.

\subsection{Reservoir Model Space (RMS)}
\label{subsec:rms}
In the RMS,
linear regression is first performed to predict the reservoir state,
one time ahead.
\begin{equation}
 \label{eq:rms1}
 \boldsymbol{x}(k+1)=R_{\mathrm{rms}}\boldsymbol{x}(k)+\boldsymbol{r}_{\mathrm{rms}}.
\end{equation}
Here, $R_{\mathrm{rms}}$ and $\boldsymbol{r}_{\mathrm{rms}}$ are
constant regardless of $k$ and are subjected to regression.
Introducing $R'_{\mathrm{rms}}$ with a definition similar to
$R'_{\mathrm{oms}}$ yields
$\boldsymbol{x}(k+1)=R'_{\mathrm{rms}}\boldsymbol{x}'(k)$.
Again, similarly to the OMS, using
 \begin{align}
 \label{rrms}
 X^+ &\equiv[\boldsymbol{x}(1), \boldsymbol{x}(2), \ldots, \boldsymbol{x}(T)] \quad \mathrm{and} \\
 X'  &\equiv[\boldsymbol{x}'(0), \boldsymbol{x}'(1), \ldots, \boldsymbol{x}'(T-1)],
\end{align}
the ridge regression of Eq.~(\ref{rrms}) leads to
\begin{equation}
 \label{eq:rms2}
 R'_{\mathrm{rms}}=X^+X'^T(X'X'^T+\lambda E)^{-1}.
\end{equation}
The RMS is a single-column reservoir representation, defined as follows~\cite{bianchi2020reservoir}:
\begin{equation}
 \boldsymbol{r} \equiv\mathrm{vec}(R'_{\mathrm{rms}}).
\end{equation}

The RMS has $N_x\cdot (N_x+1)$ features, which is
larger than the OMS.  As a result, the RMS exhibits
even higher accuracy than the OMS~\cite{bianchi2020reservoir}
at the cost of calculating the inversion of an even higher-dimensional
matrix than the OMS, which leads to a larger circuit size.

%% file: sub/propose.input_tex

\section{Delayed Feedback Reservoir with Dot-Product-based Reservoir Representation}
\label{sec:propose}

\subsection{Dot-Product-based Reservoir Representation (DPRR)}
\label{subsec:propose}

Here, we propose DPRR, which is
a novel reservoir representation based on the dot product across reservoir states of different times.

Both the OMS and RMS are effective for improving inference
accuracy~\cite{bianchi2020reservoir}.  However, the calculation of both reservoir
representations involves the inversion of a large matrix that
significantly increases the circuit size when implemented in hardware.

The fundamental requirements of reservoir representation are as follows:
1) it should retain information in a high-dimensional space;
2) its size should be constant and independent of the time series; and
3) it should be easy to calculate.

The vector of a node in the reservoir:
\begin{equation}
 \boldsymbol{x}_n\equiv[x(1)_n, x(2)_n, \ldots, x(T)_n],
\end{equation}
can be considered to be the time evolution of the node state.
A new reservoir representation can be derived by taking the dot product of the vectors of two nodes:
\begin{equation}
 \boldsymbol{x}_i\cdot\boldsymbol{x}_j = \sum_{k=1}^{T}x(k)_ix(k)_j. \;(i, j=1, 2, \ldots, N_x).
\end{equation}
This yields $N_x^2$ features that are independent of the length of the series $T$.
However, because $\boldsymbol{x}_i\cdot\boldsymbol{x}_j = \boldsymbol{x}_j\cdot\boldsymbol{x}_i$ holds,
the actual number of features is reduced by approximately half.
Furthermore, this representation completely loses information
regarding the time variation in the reservoir state.

To eliminate the loss of features while retaining the information concerning time variation, 
we  utilize the time evolution of the features with samples of shifted time.
\begin{eqnarray}
 \boldsymbol{x}_n &\equiv& [x(1)_n, x(2)_n, \ldots, x(T)_n], \\
 \boldsymbol{x}_n^- &\equiv& [x(0)_n, x(1)_n, \ldots, x(T-1)_n].
\end{eqnarray}
Taking the dot product of these vectors for two nodes yields:
\begin{equation}
 \boldsymbol{x}_i\cdot\boldsymbol{x}_j^- = \sum_{k=1}^{T}x(k)_ix(k-1)_j. \;(i, j=1, 2, \ldots, N_x).
\end{equation}
By shifting the dot product by one time step
and considering all node combinations,
we can obtain $N_x^2$ features, which are independent of the time series,
while retaining the information of the temporal evolution.
Note also that the dot product of the features at different times is not commutative.
In addition, we added the reservoir state itself as a feature:
\begin{equation}
 \boldsymbol{x}_i\cdot\textbf{1} = \sum_{k=1}^{T}x(k)_i. \;(i=1, 2, \ldots, N_x).
\end{equation}
The above $N_x\cdot (N_x+1)$ values were used as the proposed DPRR.

These features are represented as vectors. Using
$\boldsymbol{x}'(k)\equiv[\boldsymbol{x}(k),1]$,
DPRR can be represented as follows:
\begin{equation}
 \boldsymbol{r}\equiv\mathrm{vec}(\sum_{k=1}^{T}\boldsymbol{x}(k)\boldsymbol{x}'(k-1)).
\end{equation}
DPRR is expected to
be implemented more efficiently
than OMS or RMS because it can be calculated by using only matrix
multiplications.

\subsection{Fully Digital DFR for Hardware Implementation}
\label{subsec:FPGA}

Here, we describe a DFR model that uses the proposed reservoir representation.
In contrast to existing DFRs that use a mixture of digital and
analog signals, our model operates entirely digitally
making it particularly easy to implement in FPGAs.
In addition, the proposed model incorporates the following:
1) the reservoir layer applies a lightweight nonlinear function
	   without storing weights that require a large amount of memory;
2) the masking method suppresses the accuracy variation
      and supports multivariate time series inputs; and
3) the learning method of an output layer based on ridge regression with a constant term
      improves classification accuracy.

\revision{
Algorithms~\ref{alg_f} and \ref{alg_MG}
show pseudo codes for the algorithms used in the nonlinear element part of the DFR.}

\begin{algorithm}[H]
 \caption{\revision{Calculate $f$}}
 \label{alg_f}
 \begin{algorithmic}[1]
  \renewcommand{\algorithmicrequire}{\textbf{Input:}}
  \REQUIRE $x, j$
  \STATE $t \leftarrow (x+\gamma \cdot j)$
  \RETURN $\eta t/[1+t^2]$
 \end{algorithmic}
\end{algorithm}
\begin{algorithm}[H]
  \caption{\revision{Calculate {\it MackeyGlass}}}
  \label{alg_MG}
  \begin{algorithmic}[1]
  \renewcommand{\algorithmicrequire}{\textbf{Input:}}
  \REQUIRE $\boldsymbol{x}(k-1) \equiv [x(k-1)_1,x(k-1)_2,\ldots,x(k-1)_{N_x}]$
  \REQUIRE $\boldsymbol{j}(k) \equiv [j(k)_1,j(k)_2,\ldots,j(k)_{N_x}]$
  \STATE $x(k)_1 \leftarrow x(k-1)_{N_x}\mathrm{e}^{-\theta}+(1-\mathrm{e}^{-\theta})f(x(k-1)_1, j(k)_1)$
  \FOR {$n=2$ to $N$}
  \STATE $x(k)_n \leftarrow x(k)_{n-1}\mathrm{e}^{-\theta}+(1-\mathrm{e}^{-\theta})f(x(k-1)_n, j(k)_n)$
  \ENDFOR
  \RETURN $\boldsymbol{x}(k)$
  \end{algorithmic}
\end{algorithm}

\begin{figure*}[th]
 \centering
 \includegraphics[width=0.75\linewidth]{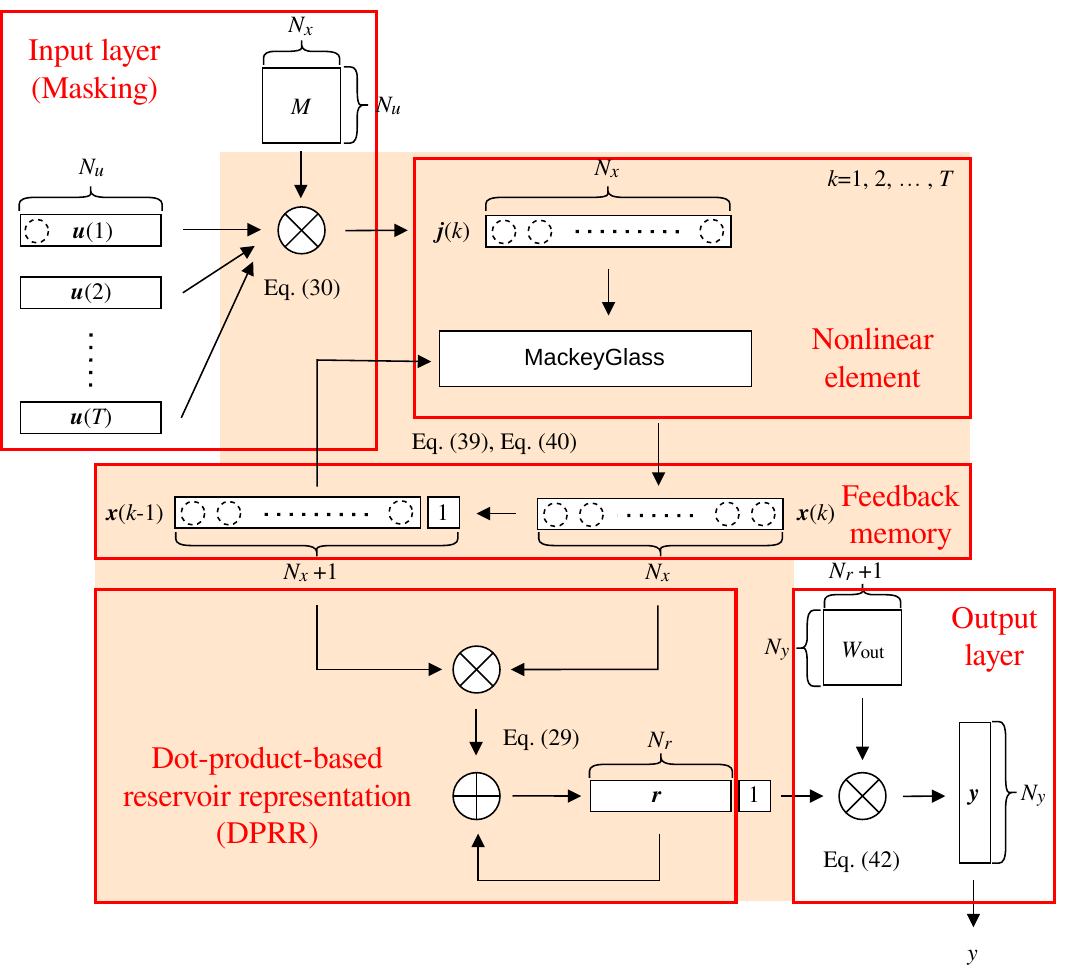}
 \caption{Circuit block diagram and dataflow of the proposed fully digital DFR with DPRR.
              \revision{The operation of "MackeyGlass" is defined in Algorithms~\ref{alg_f} and \ref{alg_MG}.}}
 \label{fig:DFRFPGA}
\end{figure*}

Figure~\ref{fig:DFRFPGA} shows a block diagram of the proposed fully
digital DFR used in this study.

The signal after the masking process $\boldsymbol{j}(k)$ is expressed as:~\cite{appeltant2011information}
\begin{equation}
 \boldsymbol{j}(k)=M\boldsymbol{u}(k).
\end{equation}
Here, $M$ is a masking matrix whose elements are either $-1$ or $1$.
The size of $M$ is $N_x\times N_u$.
The elements of $M$ can be determined randomly.
However, through separate experiments, we found that this resulted in
a larger variation in classification accuracy.
To suppress this accuracy variation associated with the randomly determined
elements, each column of $M$ is determined pseudo-randomly based on the
maximum length sequence (m-sequence)~\cite{appeltant2014constructing}.
The underlying assumption in this method is that it is desirable for the
mask function to contain various combinations of consecutive values.
For example, consider a sequence of two consecutive numbers in `01001'
--- this contains three combinations: '01', '10', and '00.' We
intended to maximize the variation in the combinations included in the
sequence.

Here, we explain how the mask functions are generated based on the m-sequence.
This was briefly explained in~\cite{appeltant2014constructing}
for the univariate case.  We extend the procedure
to multivariate time-series inputs.

The proposed procedure is summarized as follows:
\begin{enumerate}
\item Choose a primitive polynomial to use.
\item Determine an initial value that has the same length as the degree $m$ of the primitive polynomial.
\item Calculate the m-sequence.
\item Insert a zero immediately after $(m-1)$ consecutive zeros.
      If there is no such location, we insert zero at the beginning.
\item At the end of the sequence, append the $(m-1)$ initial values
      with the trailing digit removed.
\item Replace 0 with
         $-1$ to obtain a column vector of $M$.
\item For multivariate time-series inputs,
         generate column vectors of $M$
         by the round rotation of the sequence derived above.
\end{enumerate}

We show an example of
how a sequence can be generated
by choosing a primitive polynomial $G(x)=x^3+x+1$ of degree $m=3$.  In
this case, we are interested in variations in the sequence of three
consecutive numbers.  Assuming the maximal length sequence to be
$(a_n)$, the recurrence relation (base 2) can be expressed as follows:
\begin{equation}
 a_n=a_{n-2}+a_{n-3}.
\end{equation}
When the given set of initial value
is $a_0=0$, $a_1=0$, and $a_2=1$, the m-sequence is derived as follows:
\begin{equation}
 (a_n)=\{0, 0, 1, 0, 1, 1, 1\}.
\end{equation}
Among the eight different sequences of three consecutive binary numbers,
$(a_n)$ contains all sequences except `000'
when considering wrap-around rotations.
Without the rotation, $(a_n)$ does not contain `110' and `100'.
An operation is conducted to insert a zero immediately after $(m-1)$ consecutive zeros,
which yields a `000' sequence,
and $a_0, \ldots, a_{m-2}$ at the end,
which yields `110' and `100'.
\begin{equation}
 (a_n')=\{0, 0, \underline{0}, 1, 0, 1, 1, 1, \underline{0}, \underline{0}\}.
\end{equation}
After replacing the $0$ values with $-1$,
the column vector of $M$ is represented as follows:
\begin{equation}
 \{-1, -1, -1, 1, -1, -1, -1, -1, 1, 1\}.
\end{equation}
Therefore, $N_x=2^m+m-1$ holds for primitive polynomials of degree $m$.

After masking, the reservoir state $\boldsymbol{x}(k)$ is obtained
via the delayed feedback path.
As a nonlinear function,
the Mackey-\revision{Glass} model in Eq.~(\ref{eq:MG}) is
adopted.  The parameter $p=2$ is chosen in our fully digital
implementation to reduce computational resources while retaining
nonlinearity.

\revision{The secod term in Eq.~(\ref{eq:MG}) can be rewritten as $f$:
\begin{equation}
 \label{eq:f}
 f(x,t)\equiv\frac{\eta [x+\gamma j]}{1+[x+\gamma j]^p}.
\end{equation}
}
Solving the differential equation in \revision{Eq.~(\ref{eq:MG})} by assuming $f$ to be constant
for a small time $\theta$, \revision{which is the interval of virtual nodes,} yields $x(t)$:
\begin{equation}
 x(t)=x_0\e^{-t}+(1-\e^{-t})f(x(t-\tau), j(t)).
\end{equation}
Here, $x_0$ is the initial value of $x(t)$ at each $\theta$~\cite{appeltant2011information}.
\revision{Then, the value of the next virtual node of the node with $x_0$ is represented by $x(\theta)$.}
All of these operations are carried out in digital domain. Let
the components of $\boldsymbol{x}(k)$ and $\boldsymbol{j}(k)$ be:
\begin{eqnarray}
 \boldsymbol{x}(k)\; &\equiv\; [x(k)_1,x(k)_2,\ldots,x(k)_{N_x}], \\
 \boldsymbol{j}(k)\; &\equiv\; [j(k)_1,j(k)_2,\ldots,j(k)_{N_x}].
\end{eqnarray}
Then, $\boldsymbol{x}(k)$ is derived recurrently as follows:
\begin{eqnarray}
 \label{eq:digitalDFR1}
  x(k)_1 \!\!\!&=&\!\! x(k-1)_{N_x}\e^{-\theta}\nonumber\\
 &&+(1-\e^{-\theta})f(x(k-1)_1, j(k)_1), \\
 \label{eq:digitalDFR2}
 x(k)_n \!\!\!&=&\!\! x(k)_{n-1}\e^{-\theta}\nonumber\\
 &&+(1-\e^{-\theta})f(x(k-1)_n, j(k)_n).\ (n\geq2)
\end{eqnarray}
We use 0 as the initial values of the reservoir state.

The proposed DPRR uses a time-shifted feature,
wherein it is necessary to store not only $\boldsymbol{x}(k)$ but also $\boldsymbol{x}(k-1)$.
For this purpose, the memory required to store $\boldsymbol{x}(k)$ is duplicated.
\revision{As $\theta$ is constant,} the values of $\e^{-\theta}$ and $(1-\e^{-\theta})$ are
constants and they can therefore be precalculated and embedded in the logic
circuit.

The recurrent equation approach for solving
differential equations is intended to balance the
accuracy and computational complexity.
In a previous study, either the Euler~\cite{alomar2015digital} or Heun~\cite{appeltant2014constructing}
methods were used.
The Euler method is relatively simple but
may compromise the classification accuracy,
whereas the Heun method is relatively complex and requires a larger circuit for
hardware implementation.

Next, the reservoir representation $\boldsymbol{r}$
is obtained using DPRR from the reservoir states.

Finally, output $\boldsymbol{y}$ is obtained.
First, a constant term is added to improve the classification accuracy:
\begin{equation}
 \boldsymbol{r}'\equiv[\boldsymbol{r}, 1].
\end{equation}
The weight matrix $W_{\mathrm{out}}$ of size $N_y\times (N_r+1)$
is used to compute $\boldsymbol{y}$:
\begin{equation}
 \boldsymbol{y} = W_{\mathrm{out}}\boldsymbol{r}'.
\end{equation}
The largest value in the elements of $\boldsymbol{y}$ is used as the inference label.
Ridge regression is used to determine $W_{\mathrm{out}}$.

%% file: sub/evaluation.input_tex

\section{Evaluation}
\label{sec:evaluation}

\revision{In this section we evaluate the classification accuracy
and required hardware resources of the proposed fully digital DFR with
DPRR.  First, a comparison with different reservoir representations is
made.
Thereafter, we compare our results with those of other machine learning methods.
Comparisons are made through both software and FPGA implementations, the
former to evaluate classification performance and the latter to evaluate
hardware resource usage.  Finally, the tradeoffs between analog and full
digital implementation are discussed.}

\revision{For the FPGA implementation, we used Xilinx Vitis HLS 2021.1
as the high-level synthesis tool and Zynq-7000
(xc7z020clg400-1) as the target FPGA board.
The numbers are represented in IEEE-754 floating-point format, and the data
paths were not pipelined.
In practice, it may be possible to reduce the
bit width of number representations and to improve
throughput by pipelining.  However,
as it is difficult to
ensure an equal level of optimizations across multiple machine
learning methods, this setting was selected
for fair comparison.}

A summary of the datasets used in the evaluation is listed in Table~\ref{tab:dataset}.
Throughout the evaluation, we used the npz files available in~\cite{bianchi2020reservoir}.
\begin{table}
 \centering
 \caption{Summary of multivariate time series classification datasets \revision{(taken from \cite{bianchi2020reservoir}).
 Column \#$V$, \#$C$, Train, and Test respectively reports
 the input dimension, the output dimension, the number of training
 data, and that of testing data.  $T_\mathrm{min}$ and
 $T_\mathrm{max}$ are the shortest and the longest length of the input
 series in the dataset, respectively.}}
 \label{tab:dataset}
 \begin{tabular}{c|cccccc} \hline
  Dataset & \#$V$ & \#$C$ & Train & Test & $T_{\mathrm{min}}$ & $T_{\mathrm{max}}$\\ \hline \hline
  ARAB & 13 & 10 & 6600 & 2200 & 4 & 93 \\ \hline
  AUS & 22 & 95 & 1140 & 1425 & 45 & 136 \\ \hline
  CHAR & 3 & 20 & 300 & 2558 & 109 & 205 \\ \hline
  CMU & 62 & 2 & 29 & 29 & 127 & 580 \\ \hline
  ECG & 2 & 2 & 100 & 100 & 39 & 152 \\ \hline
  JPVOW & 12 & 9 & 270 & 370 & 7 & 29 \\ \hline
  KICK & 62 & 2 & 16 & 10 & 274 & 841 \\ \hline
  LIB & 2 & 15 & 180 & 180 & 45 & 45 \\ \hline
  NET & 4 & 13 & 803 & 534 & 50 & 994 \\ \hline
  UWAV & 3 & 8 & 200 & 427 & 315 & 315 \\ \hline
  WAF & 6 & 2 & 298 & 896 & 104 & 198 \\ \hline
  WALK & 62 & 2 & 28 & 16 & 128 & 1918 \\ \hline
 \end{tabular}
\end{table}

\subsection{Effects of Reservoir Representation}
\label{subsec:comp1}

First, the effect of reservoir representation was studied.  DFR with
various reservoir representations was implemented in Python.
The dataset used in this section was \revision{ARAB}.

Because different reservoir representations yield different value ranges,
the hyperparameters $\gamma$ and $\eta$ must be adjusted to attain
improved accuracy. Table~\ref{tab:parameter_comp1} shows
the hyperparameters used in this evaluation that were found through a grid search
in the range \{0.03, 0.1, 0.3, 1\}. Here,
MRS\_upad and MRS\_xpad are
the maximal reservoir states
with zero-filling for the input and reservoir states, respectively.
\begin{table}[th]
 \centering
 \caption{Parameters used in comparison with existing reservoir representation.
 The parameters
 \revision{$\theta=0.25$, $\beta=0.01$, and $\lambda=1$ are commonly used. The dataset is ARAB.}}
 \label{tab:parameter_comp1}
 \begin{tabular}{c|ccccccc} \hline
  & LRS & DRS & MRS & MRS & OMS & RMS & DPRR \\
  & & & \_upad & \_xpad & & & \\ \hline \hline
  $\gamma$ & 0.03 & 0.3 & 0.1 & 0.03 & 0.03 & 0.03 & 0.03 \\ \hline
  $\eta$ & 1 & 0.1 & 0.1 & 1 & 1 & 1 & 1 \\ \hline
 \end{tabular}
\end{table}

Table~\ref{tab:acc_comp1} lists the classification accuracy
for each reservoir representation.
\begin{table}[tbh]
 \centering
 \caption{Classification accuracy for different reservoir representation.
 \revision{The dataset used is ARAB.
 Classification accuracy is in \%.
 $m = 3, 4, 5, 6$ corresponds to $N_x = 10, 19, 36, 69$, respectively.}}
\label{tab:acc_comp1}
\begin{tabular}{c|ccccccc} \hline
 & LRS & DRS & MRS & MRS & OMS & RMS & DPRR \\
$m$ & & &\_upad &\_xpad & & & \\ \hline \hline
$3$ & 44.9 & 18.2 & 89.3 & 89.1 & 90.9 & 85.4 & 88.5 \\ \hline
$4$ & 52.6 & 22.0 & 93.0 & 92.6 & 95.3 & 95.0 & 96.5 \\ \hline
$5$ & 54.1 & 26.0 & 93.5 & 93.2 & 96.2 & 96.0 & 97.5 \\ \hline
$6$ & 60.1 & 38.1 & 94.3 & 94.7 & 96.0 & 98.0 & 98.0 \\ \hline
\end{tabular}
\end{table}

Regardless of the choice of $m$,
LRS and DRS were less accurate than the other reservoir representations.
Although the proposed DPRR is computationally efficient,
it achieved equivalent or higher accuracy compared to OMS and RMS.

\revision{The calculation of each reservoir representation transformation was implemented
on an FPGA using a high-level synthesis tool to demonstrate
that DPRR is computationally more efficient than OMS or RMS.
Table~\ref{tab:synth_r} presents the logic synthesis results.}
\begin{table}[tpb]
 \centering
 \caption{\revision{Synthesis results for different
		reservoir representation. Latency is in Cycles.
		Based on the ARAB dataset, the dimension of the input is 13 and the series length is 50.}}
 \label{tab:synth_r}
\begin{tabular}{cc|rrrrr} \hline
& \multicolumn{1}{r}{$N_x$} & \multicolumn{1}{c}{10} & \multicolumn{1}{c}{20} & \multicolumn{1}{c}{30} & \multicolumn{1}{c}{40} & \multicolumn{1}{c}{50} \\
\hline \hline
& Latency & 16,344 & 62,032 & 162,125 & 412,452 & 740,512 \\
& BRAM & 8 & 8 & 11 & 21 & 37 \\
OMS & DSP & 18 & 18 & 12 & 19 & 19 \\
& FF & 7,822 & 10,941 & 15,422 & 12,529 & 14,325 \\
& LUT & 8,631 & 11,080 & 13,490 & 11,522 & 12,761 \\ \hline
& Latency & 14,474 & 63,649 & 170,557 & 435,699 & 789,574 \\
& BRAM & 8 & 8 & 12 & 26 & 50 \\
RMS & DSP & 19 & 19 & 12 & 20 & 19 \\
& FF & 7,768 & 10,900 & 15,455 & 12,510 & 14,351 \\
& LUT & 8,284 & 10,615 & 12,805 & 10,643 & 11,734 \\ \hline
& Latency & 5,628 & 21,448 & 47,468 & 83,688 & 130,108 \\
& BRAM & 2 & 2 & 2 & 5 & 9 \\
DPRR & DSP & 5 & 5 & 5 & 5 & 5 \\
& FF & 1,212 & 1,815 & 2,284 & 2,667 & 3,171 \\
& LUT & 1,465 & 1,836 & 2,086 & 2,395 & 2,725 \\ \hline
\end{tabular}
\end{table}

\revision{Clearly, DPRR is more computationally efficient than OMS and RMS.
When $N_x=50$, the number of FFs in DPRR is reduced to
about 1/4 of those in OMS and in RMS, and the number
of LUTs is about 1/5.}

\subsection{Comparison of Classification Accuracy with Existing Machine Learning Methods}
\label{subsec:comp2}

Next, we compared the accuracy of the proposed DFR with DPRR
(DFR\_DPRR), to those of the DFR using DRS (DFR\_DRS)
and other machine learning methods~\cite{IsmailFawaz2018deep}.
\revision{Comparison with DFR\_DRS corresponds to comparison with existing studies of DFR~\cite{appeltant2011information,soriano2014delay}.
This comparison of accuracy is followed by a comparison of hardware usage.
As can be seen from the comparison in Section~\ref{subsec:comp1}, DFR with OMS or RMS has the same level of accuracy as DFR\_DPRR,
but is less computationally efficient than DFR\_DPRR,
therefore it is not included in the comparison.
However, DFR\_DRS is less accurate
than DFR\_DPRR, but is more computationally efficient,
therefore it is worth evaluating.}
The DFR models were implemented using Python.
The hyperparameters
listed in Table~\ref{tab:parameter} were used for each dataset. 
\begin{table}[thb]
 \centering
 \caption{Parameters of the DFR \revision{used for comparision with existing machine learning methods.
	The number of nodes $N_x$ in the reservoir is 36 ($m=5$).}}
 \label{tab:parameter}
 \begin{tabular}{c|cccc|cccc} \hline
  & \multicolumn{4}{c|}{DFR\_DRS} & \multicolumn{4}{c}{DFR\_DPRR} \\
\revision{Dataset}  & $\gamma$ & $\eta$ & $\theta$ & $\beta$ & $\gamma$ & $\eta$ & $\theta$ & $\beta$ \\
\hline \hline
ARAB & 4 & 0.4 & 0.3 & 1e-2 & 0.04 & 1 & 0.3 & 1e-2 \\ \hline
AUS & 0.3 & 4 & 0.45 & 1e-1 & 0.05 & 1 & 0.25 & 1e-3 \\ \hline
CHAR & 0.2 & 1.5 & 0.4 & 1e-2 & 0.05 & 1 & 0.3 & 1e-4 \\ \hline
CMU & 0.02 & 0.1 & 0.2 & 1e-2 & 0.01 & 3 & 0.15 & 1e-1 \\ \hline
ECG & 1 & 1 & 0.15 & 1e-1 & 0.03 & 1 & 0.3 & 1e-5 \\ \hline
JPVOW & 0.1 & 1 & 0.2 & 1e-2 & 0.03 & 1 & 0.2 & 1e-1 \\ \hline
KICK & 3 & 3 & 0.15 & 1e-1 & 0.003 & 3 & 0.15 & 1e-1 \\ \hline
LIB & 0.3 & 1.5 & 0.25 & 1e-5 & 0.14 & 1 & 0.45 & 1e-3 \\ \hline
NET & 0.01 & 3 & 0.2 & 1e-2 & 0.025 & 3 & 0.2 & 1e-3 \\ \hline
UWAV & 0.03 & 1.5 & 0.15 & 1e-5 & 0.03 & 1 & 0.4 & 1e-2 \\ \hline
WAF & 1 & 1 & 0.15 & 1e-3 & 0.2 & 0.4 & 0.25 & 1e-2 \\ \hline
WALK & 0.01 & 1 & 0.15 & 1e-1 & 1 & 1 & 0.25 & 1e-2 \\ \hline
 \end{tabular}
\end{table}

In some datasets, the length of the input series varied.
For data with a 
shorter length $T$ than the maximum series length
$T_{\mathrm{max}}$, the inputs were filled with zeros when using
methods other than DFRs.

The machine learning methods that were compared were:
\begin{enumerate}
\item Multi layer perceptron (MLP) \cite{wang2017time}
\item Fully convolutional neural network (FCN) \cite{wang2017time}
\item Residual network (ResNet) \cite{wang2017time}
\item Encoder \cite{serra2018towards}
\item Multi channel deep convolutional neural network (MCDCNN) \cite{zheng2014time}
\item Time convolutional neural network (Time-CNN) \cite{zhao2017convolutional}
\item Time warping invariant echo state network (TWIESN). \cite{tanisaro2016time}
\end{enumerate}

\revision{In~\cite{IsmailFawaz2018deep},
MCNN and t-LeNet were also evaluated.
However, these two were inferior to the above methods in terms of Average Rank, which will be discussed later, and
have been omitted for clarity of comparison.}

Table~\ref{tab:acc} summarizes the classification accuracies of all the aforementioned methods.
\begin{table*}
\centering
 \caption{Classification accuracy with other machine learning methods.
 \revision{Accuracy is in \%. The best and the second-best accuracies for each dataset is shown in bold and
 underlined respectively.
 The average accuracy is defined as the arithmetic mean of the accuracy.
 The average rank is defined as the arithmetic mean of the rank numbers being the most accurate method as 1.
 The data in the table, except for DFR, are from~\cite{IsmailFawaz2018deep}.}}
\label{tab:acc}
\begin{tabular}{c|ccccccccc} \hline
& MLP & FCN & Res & Encoder & MCD & Time- & TWI & DFR\_ & DFR\_ \\
\revision{Dataset} &  &  & Net &  & CNN & CNN & ESN & DRS & DPRR \\
\hline \hline
ARAB & 96.9 & \underline{99.4} & \bf{99.6} & 98.1 & 95.9 & 95.8 & 85.3 & 27.8 & 98.0 \\ \hline
AUS & 93.3 & \bf{97.5} & \underline{97.4} & 93.8 & 85.4 & 72.6 & 72.4 & {\revision{34.4}} & 95.6 \\ \hline
CHAR & 96.9 & \bf{99.0} & \bf{99.0} & 97.1 & 93.8 & 96.0 & 92.0 & 33.9 & 96.2 \\ \hline
CMU & 60.0 & \bf{100.0} & 99.7 & 98.3 & 51.4 & 97.6 & 89.3 & 93.1 & \bf{100.0} \\ \hline
ECG & 74.8 & \underline{87.2} & 86.7 & \underline{87.2} & 50.0 & 84.1 & 73.7 & 67.0 & \bf{88.0} \\ \hline
JPVOW & 97.6 & \bf{99.3} & \underline{99.2} & 97.6 & 94.4 & 95.6 & 96.5 & 62.4 & 97.8 \\ \hline
KICK & 61.0 & 54.0 & 51.0 & 61.0 & 56.0 & 62.0 & \underline{67.0} & 60.0 & \bf{80.0} \\ \hline
LIB & 78.0 & \bf{96.4} & \underline{95.4} & 78.3 & 65.1 & 63.7 & 79.4 & 30.0 & 78.3 \\ \hline
NET & 55.0 & 89.1 & 62.7 & 77.7 & 63.0 & 89.0 & \underline{94.5} & 92.5 & \bf{95.9} \\ \hline
UWAV & 90.1 & \bf{93.4} & \underline{92.6} & 90.8 & 84.5 & 85.9 & 75.4 & 26.2 & 86.2 \\ \hline
WAF & 89.4 & 98.2 & \underline{98.9} & 98.6 & 65.8 & 94.8 & 94.9 & 90.2 & \bf{99.0} \\ \hline
WALK & 70.0 & \bf{100.0} & \bf{100.0} & \bf{100.0} & 45.0 & \bf{100.0} & 94.4 & \bf{100.0} & \bf{100.0} \\ \hline \hline
\revision{Average Accuracy} & \revision{80.3} & \revision{92.8} & \revision{90.2} & \revision{89.9} & \revision{70.9} & \revision{86.4} & \revision{84.6} & \revision{59.8} & \revision{92.9} \\ \hline
Average Rank & 5.92 & 2.25 & 3.08 & 3.42 & 7.58 & 5.50 & 5.92 & 7.08 & 2.50 \\ \hline
\end{tabular}
\end{table*}
The proposed method ranks second to the FCN in terms of average rank
and consistently exhibits high accuracy on all datasets.  In terms of the
classification accuracy, the proposed method is comparable to FCN
and outperforms most other machine learning methods.

\subsection{Comparison of Hardware Resource Usage with Existing Machine Learning Methods}
\label{subsec:comp3}

The proposed model was implemented on an FPGA using a high-level
synthesis tool to investigate its hardware resource usage.
\revision{The Python code used in Section~\ref{subsec:comp1} for DFR and
the Python code available in~\cite{IsmailFawaz2018deep} for the other
methods were rewritten in C++ and synthesized for the target
FPGA.}  Table~\ref{tab:synth} presents the logic synthesis results.
\begin{table}[tpb]
 \centering
 \caption{\revision{Synthesis results for different
 machine learning methods.
 Latency is in cycles. The dataset used is ARAB.}}
 \label{tab:synth}
\begin{tabular}{c|rrrrr} \hline
\revision{Method} & \multicolumn{1}{c}{Latency} & \multicolumn{1}{c}{BRAM} & \multicolumn{1}{c}{DSP} & \multicolumn{1}{c}{FF} & \multicolumn{1}{c}{LUT} \\
\hline \hline
MLP & 6,663,799 & 3,105 & 9 & 3,784 & 4,575 \\ \hline
FCN & 126,841,718 & 942 & 12 & 5,277 & 7,383 \\ \hline
ResNet & 227,560,415 & 1,951 & 87 & 34,673 & 37,984 \\ \hline
Encoder & 801,121,336 & 5,787 & 19 & 9,941 & 12,395 \\ \hline
MCDCNN & 16,328,669 & 8,547 & 62 & 11,909 & 14,882 \\ \hline
Time-CNN & 270,020 & 61 & 74 & 14,488 & 16,600 \\ \hline
TWIESN & 6,229,486 & 330 & 24 & 3,529 & 4,946 \\ \hline
DFR\_DRS & 2,130 & 21 & 182 & 22,905 & 30,767 \\ \hline
DFR\_DPRR & 84,532 & 57 & 65 & 12,083 & 14,152 \\ \hline
\end{tabular}
\end{table}

\begin{figure}[thb]
 \centering
 \includegraphics[width=1.0\linewidth]{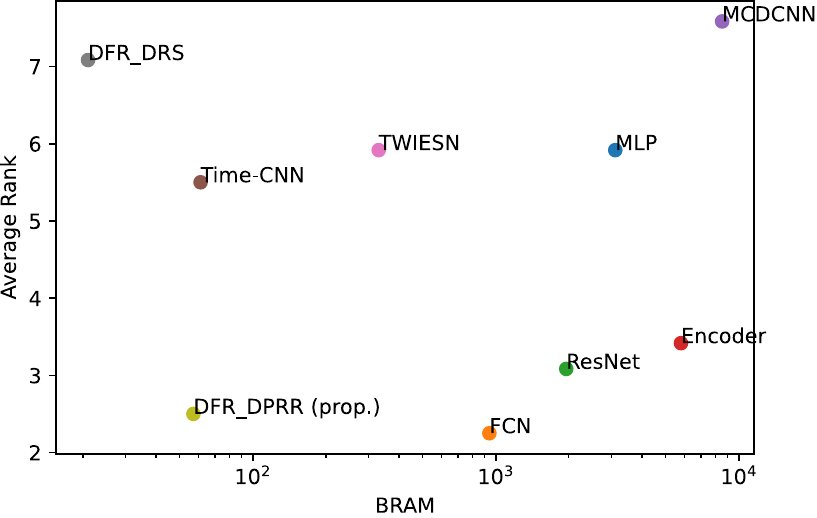}
 \caption{\revision{Comparison of memory usage. Each method is plotted using BRAM counts on the horizontal axis and
  the average rank calculated in Section~\ref{subsec:comp2} on the vertical axis}}
 \label{fig:BRAM}
\end{figure}

Figure~\ref{fig:BRAM} compares the number of block RAMs
(BRAMs) required for the FPGA implementation.
The smaller the numbers on both the horizontal and vertical axes
indicate improved results (i.e., methods with results closer to the
lower left achieves high accuracy with a smaller memory footprint).
Methods based on DFR require a much smaller memory footprint than most
machine-learning methods and ESN.  Only Time-CNN with a very small
number of filters in the convolution layers consumes approximately an
equal number of BRAMs, but the accuracy of the proposed method is
significantly better than that of Time-CNN.
\revision{If DFR with OMS or RMS were plotted in
the same figure, they would be located directly to the right
of DFR\_DPRR, i.e., similar accuracy while about 5x larger BRAM
size.}

\revision{The number of BRAMs on the target FPGA board (Zynq-7000,
xc7z020clg400-1) is 220, and thus the methods except DFR and
Time-CNN cannot be implemented.  Even if Zynq UltraScale+ MPSoC ZCU102,
were used, in which the total number of available
BRAMs is 892 (32.1\,Mb),
the other methods except for TWIESN still could not be implemented.
Therefore, when comparing the hardware implementation
efficiency using the power-delay product, which is equivalent to energy consumption,
the BRAMs were excluded and the power consumption of other elements
(i.e., LUT, FF, and DSP) was calculated using Xilinx power estimator
(XPE).  XPE version 2019.1.2 is used for Zynq-7000 as target.
"Quick Estimate" was used to calculate the power and the setting was set
as default except for "Design Utilization".}
\revision{Table~\ref{tab:estimation} presents the summary of power
estimation results.}
\begin{table}[tpb]
 \centering
 \caption{\revision{Power estimation results. Power is in W.
			For each method, the resource usage obtained from the high-level synthesis was entered in "Design Utilization".		
			Note that these powers are independent of the number of BRAMs.}}
 \label{tab:estimation}
\begin{tabular}{c|rrr|r} \hline
Method & \multicolumn{1}{c}{CLOCK} & \multicolumn{1}{c}{LOGIC} & \multicolumn{1}{c}{DSP} & \multicolumn{1}{|c}{SUM}\\
\hline \hline
MLP & 0.102 & 0.095 & 0.004 & 0.201 \\ \hline
FCN & 0.111 & 0.110 & 0.005 & 0.226 \\ \hline
ResNet & 0.254 & 0.313 & 0.038 & 0.605 \\ \hline
Encoder & 0.137 & 0.143 & 0.008 & 0.288 \\ \hline
MCDCNN & 0.147 & 0.159 & 0.027 & 0.333 \\ \hline
Time-CNN & 0.160 & 0.172 & 0.033 & 0.365 \\ \hline
TWIESN & 0.102 & 0.096 & 0.011 & 0.209 \\ \hline
DFR\_DRS & 0.202 & 0.253 & 0.080 & 0.535 \\ \hline
DFR\_DPRR & 0.148 & 0.156 & 0.029 & 0.333 \\ \hline
\end{tabular}
\end{table}

\begin{figure}[tbh]
 \centering
 \includegraphics[width=1.0\linewidth]{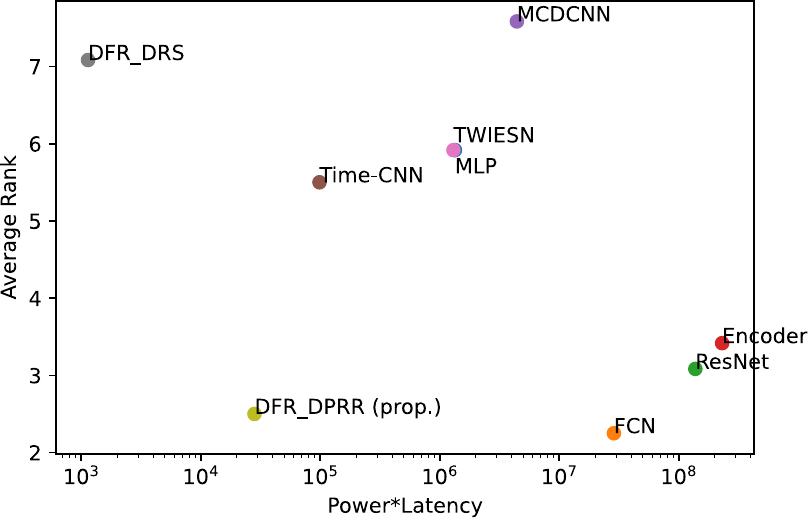}
 \caption{Accuracy versus power-delay product.
 \revision{Each method is plotted using the product of
 power consumption except BRAM and latency in clock cycles.
 The power-delay product is on the horizontal axis,
 and the average rank is on the vertical axis.}}
 \label{fig:PowerLatency}
\end{figure}

\revision{Figure~\ref{fig:PowerLatency} compares the power-delay product
using the number of delay cycles obtained from the high-level synthesis
tool and the power consumption without BRAM obtained
using the power estimation tool.}
\revision{Similar to Fig.~\ref{fig:BRAM}, the closer the results are to
the lower left, the better is the method.  The proposed method,
DFR\_DPRR, operates with a power-delay product that is three orders of
magnitude smaller than that of FCN but achieves nearly the same accuracy
as FCN. The difference in
power-delay product would become larger
if the power consumption of BRAM is considered.}

Overall, the proposed method exhibited the best balance in terms of
accuracy and hardware efficiency among the investigated methods.

\subsection{\revision{Tradeoffs between Analog and Full Digital Implementation}}
\label{subsec:comp4}

\revision{Analog and digital implementations exhibit pros and cons.
Here, we discuss the tradeoff between the analog
implementation of DFR~\cite{appeltant2011information,soriano2014delay}
and the proposed full digital implementation.}

\begin{table}[tpb]
 \centering
 \caption{\revision{Implementation results of  proposed DFR\_DPRR. The dataset used is ARAB. All 2200 testing data are processed.
		Power is the sum of the power of clocks, signals, logic, BRAM, and DSP.}}
 \label{tab:implementation}
\begin{tabular}{c|c} \hline
LUT & 12,471 (23.44 \%) \\
LUTRAM & 390 (2.24 \%) \\
FF & 17,009 (15.99 \%) \\
BRAM & 26 (18.57 \%) \\
DSP & 67 (30.45 \%) \\
BUFG & 1 (3.13 \%) \\
Clock & 100 MHz \\
Worst Negative Slack & 0.566 ns \\
Power & 0.274 W \\
Prediction Time (2200 time-series) & 7.68 s \\
Energy Consumption (/ 1 time-series) & 9.57E-04 J \\ \hline
\end{tabular}
\end{table}

\revision{An analog implementation of the reservoir would be beneficial in
terms of power or energy.  However,
neither power consumption nor energy consumption was evaluated in
existing studies~\cite{appeltant2011information,soriano2014delay}.
In general, analog implementation of DFRs require a DAC and
ADC, and the non-linear (such as Mackey-Glass) circuit
should involve the design of an operational amplifier, making
the overall circuit size larger.  In addition, the design of such
analog circuits and their components may require expertise and tuning,
which can make it difficult to achieve maximum accuracy.
The circuit design process can take a long time to complete.
In contrast, the digital implementations, particularly the proposed fully
digital DFR, are easy to adopt and more feasible with existing design
tools.
With digital design it is
also easy to adjust the size and
accuracy of the circuit by changing hyperparameters and data
representations, such as input and output dimensions and the
precision of fractional numbers.
Furthermore, as our experiments in this section indicate,
DFR\_DPRR can achieve comparable or higher accuracy
in comparison to the cutting-edge neural network-based methods
as well as analog implementations.}

\revision{Finally, to allow future quantitative comparison of design
tradeoffs among analog, digital, and other design styles, the
implementation result of the proposed DFR\_DPRR generated using Xilinx
Vivado 2021.1 is summarized in
Table~\ref{tab:implementation}.
We confirmed that the proposed DFR\_DPRR achieves
exactly the same classification accuracy obtained in the Python
simulations performed in Section~\ref{subsec:comp2}.}

%% file: sub/conclusion.input_tex

\section{Conclusion}
\label{sec:conclusion}

In this paper, we proposed a computationally efficient reservoir representation,
namely DPRR, based on
the dot product between features in the reservoir of shifted time.
The proposed DPRR contributes to
the implementation of accurate and compact hardware for reservoir-based
classification.
We also proposed a fully digital
DFR using DPRR
that is suitable
for solving multivariate time-series classification tasks, which
have been difficult to achieve high accuracy
using reservoir computing methods.
Through the FPGA implementation of the proposed DFR using DPRR,
the proposed method
was evaluated on 12 different multivariate time-series classification tasks.
The DFR using DPRR outperformed
conventional machine learning methods in terms of accuracy,
despite its hardware resources in terms of
\revision{power-delay product}
and memory usage
being orders of magnitude smaller.
The proposed model is therefore the most suitable
among all investigated methods
for solving multivariate time-series classification tasks via hardware.